\documentclass{sig-alternate}
\usepackage[utf8]{inputenc}  
\usepackage{xcolor}

\begin{document}

\setcopyright{acmlicensed}

\doi{http://dx.doi.org/10.1145/3077286.3077293}

\isbn{978-1-4503-5024-2/17/04}

\conferenceinfo{ACM SE '17}{April 13--15, 2017, Kennesaw, GA, USA}

\acmPrice{\$15.00}

%

\title{Acquisition and use of knowledge over a restricted domain by intelligent agents\titlenote{(Produces the permission block, and
copyright information). For use with
SIG-ALTERNATE.CLS. Supported by ACM.}}
%
%
%
%
%

\numberofauthors{3}
\author{
%
%
\alignauthor
Juliao Braga\\
       \affaddr{Mackenzie University}\\
       \affaddr{Rua Consolacao, 930}\\
       \affaddr{Sao Paulo, SP, Brazil}\\
       \email{juliao@braga.net.br}
\alignauthor 
Nizam Omar\\
       \affaddr{Mackenzie University}\\
       \affaddr{Rua Consolacao, 930}\\
       \affaddr{Sao Paulo, SP, Brazil}\\
       \email{nizam.omar@mackenzie.br}
\and  
\alignauthor
Luciana F. Thome\\
       \affaddr{Fluminense Federal University}\\
       \affaddr{Av. Gal. Milton T. Souza,  s/n}\\
       \affaddr{Niterói, RJ, Brazil}\\
       \email{lthome@ic.uff.br}
}
\date{23 March 2017}

\maketitle

\begin{abstract}
This short paper provides a description of an architecture to acquisition and use of knowledge by intelligent agents over a restricted domain of the Internet  Infrastructure. The proposed architecture is added to an intelligent agent deployment model over a very useful server for Internet Autonomous System administrators. Such servers, which are heavily dependent on arbitrary and eventual updates of human beings, become unreliable. This is a position paper that proposes three research questions that are still in progress.
\end{abstract}

%
%


%
%

%
%



\section{Introduction}
\label{sec:firstpage}

Paul Horn, in 2001, inspired by the living system physiology presented an IBM proposal for the future of computer systems \cite{Horn:2001}. His work argued that the efforts of specialists in the maintenance, control and operation of computer systems could be minimized and consequently have their costs reduced dramatically. The community, composed mainly of researchers continued to advance in the researches of this knowledge domain becoming a paradigm named by Horn as \textbf{Autonomic Computation}. 

Contributions have been expanded by multidisciplinary research groups and the results have been surprising \cite{Movahedi:2012}. A number of applications, particularly in software, have enabled, for example, the technology of space probes \cite{Sterritt:2005b}, rather Unmanned Space Vehicles (USVs) \cite{Insaurralde:2015}.

The interest aroused has led to the application of Intelligent Agents or Intelligent Elements (IEs) in the Infrastructure of the Internet, concentrating basically on the protocols and techniques like Software Defined Networking (SDN) \cite{Shukla:2014, Nadeau:2013, Wickboldt:2015} will encourage the development of new initiatives in this direction. It is important to develop case studies and experiments on the entire spectrum of applications for the Internet Infrastructure. In this sense, the renewed experience and expansion of research groups will make an effective contribution to improving and consolidating the studies that are being carried out to date, especially the principle of interdisciplinary cooperation.

Section 2 presents the abstract IEs model and the respective application domain. From the abstract model this section continues to the implementation model with its physical characteristics. Section 3 describes the architecture for knowledge acquisition and its functional components. Section 4 presents the steps involved in the process of knowledge acquisition and the research questions involved. Section 5 presents the conclusions with the identification of future problems. Section 6 concludes the authors' acknowledgments.

\section{The Application Domain and its characteristics}

\begin{figure*}[ht]
  \centering
  \includegraphics[width=.9\textwidth]{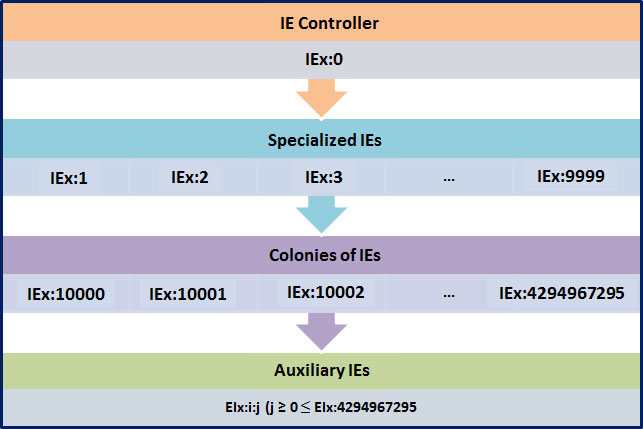}
  \caption{Four Layer Abstract Model of Autonomous  Architecture  for Restricted Domains (A2RD). Source: \cite{Braga:2015}}
  \label{fig:ModeloAbstrato_EN}
  \end{figure*}

The Autonomous Architecture for Restricted $\quad$Domains  (A2RD) model is presented in Figure \ref{fig:ModeloAbstrato_EN} and divided into four layers, described below. The model serves the interest of establishing an architecture of intelligent elements under the administrative domain of ASs, which is known as the designation given to the networks that form the Internet.

 The model can exist in any of the $2^{32}$ possible ASs \cite{Hawkinson:1996}. However, on 02/03/2017 there were 56,710  active ASs on the Internet (originating traffic), according to CIDR-report\footnote{http://www.cidr-report.org/as2.0/}. The number of an AS is unique, controlled by the Regional Internet Registers (RIRs) and / or National Internet Registers (NIRs) and is called the Autonomous System Number (ASN). Thus, the largest possible value of x is 56710, corresponding to AS56710, at the date above. There is no conflict between the model being deployed in any AS environment and being domain-restricted. In fact, the implementations are independent, but with a high degree of interoperability and, of course, intense cooperation, because ASes administrators depend on the behavior of all the others. The IANA has reserved two contiguous ranges of ASs numbers for private use \cite{Mitchell:2013}: 64512-65534 and 4200000000-4294967294. Conveniently, these ASes numbers can be used to designate Intelligent Elements in applications that need to represent subdomains.

The first of the four layers hosts the Intelligent Element (IE) named \textbf{Controller}. Its identification is unique and definitive: \textbf{x:0}, that is, the number \textbf{0} placed to the right side of the symbol \textbf{:}, following by the ASN that hosting the model. Sometimes, to make clear which IE is being referenced, \textbf{IE} is used before the identification, as for example, when affirming that the IE Controller is \textbf{IEx:0}. Thus, if \textbf{AS5} is the host domain of the model, then the controlling element is \textbf{IE5:0}. No IE of the lower layers can exist, without the prior consent of the IE Controller. It has the property of keeping oneself organized (self-organization) and ensuring the self-organization of any IE of the lower layers.

The second layer is represented by the so-called Specialized IEs. These elements are identified by suffixes that can range from \textbf{1} to \textbf{9999}. The specialized elements support the IE Controller, in specific activities and necessary to the respective functionalities. These activities range from ensuring the interoperability of the entire system of implemented IEs, to the establishment of specific functionalities, such as servers with end-to-end characteristics \cite{Saltzer:1984}, database access functionalities and semantic repositories, proprietary software (similar to Southern SDN APIs) \cite{Wickboldt:2015}, features required for lower-layer IEs. However, support for the IE Controller is the primary objective of the Specialized IEs. This objective is who determines the functionalities of the second layer. It is assumed that some Specialized IEs may be Autonomic Elements or intelligent elements that execute automatic processes, such as proprietary software and procedures associated with legacy systems, among others. A Specialized IE can be created as functions that only concern the IE Controller, especially when it depends on the functionalities of IEs of the third layer.

In the third layer lies the largest IEs agglomeration, which is why it is called the \textbf{IE Colonies}. Elements of this layer can be Autonomous, Autonomic or Automatic, except Legacy and are directly responsible for the most important activities of the application, including software reuse. They act under the influence of a high degree of interoperability and cooperation between them and between IEs of other layers and other domains / subdomains. They do not directly participate in interconnections or exchange messages with other IEs outside the domain, but they do so through the IEs of the upper layers. There is intense semantic interoperability activity by these IEs, which have a high capacity for self-learning due to continuous interactions with the domain environment, and produce improvement effects on the knowledge of other IEs of the colony itself and the IEs of the upper layers, IE Controller. In other words, these IEs favor the learning of the entire cluster of IEs of the layer model, which hour is being described. The IEs of the colonies receive an identification with numeric suffixes, ranging from \textbf{10000} to \textbf{4294967295}.

In the fourth layer are the \textbf{Auxiliary IEs}. This layer exists, in order to allow the transfer of computing demands to a new set of IEs (successiveness of the model). It reproduces, successively, the first, second, third and a new fourth layers. This new IEs sequence has an additional suffix \textbf{:j:0} for a new IE Controller responsible for the following four new layers. In the new second, third and fourth layers, the IDs of the IEs are postfixed with \textbf{:j:id}, where \textbf{j} is a colony IE number that originated the new fourth layer and the \textbf{id} is a number with the above specifications. A typical application for the fourth layer are subdomains, such as home networks (homenet).

\begin{figure}[ht]
  \centering
  \includegraphics[width=.46\textwidth]{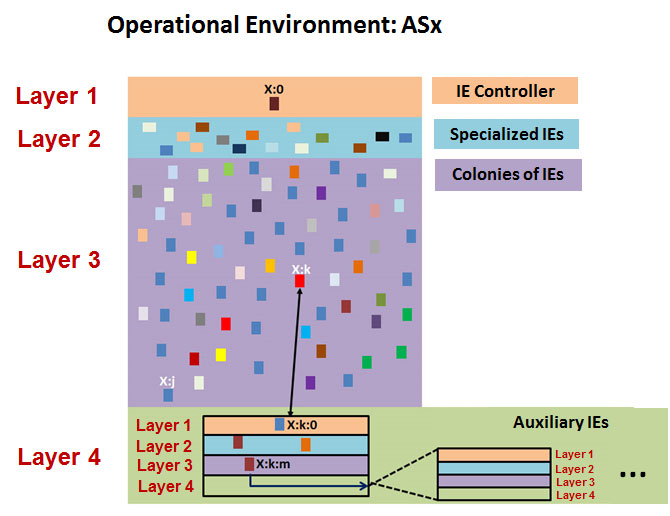}
  \caption{A2RD Implementation Model. Source: \cite{Braga:2015a} \cite{Braga:2015}}
  \label{fig:A2RD-ModeloImplementacao_EN}
  \end{figure}

The figure \ref{fig:A2RD-ModeloImplementacao_EN} is the A2RD implementation model, where the small and colored rectangles are IEs. It is seen that the IEs are arranged and distributed among the layers, similar to what was said earlier about the abstract model. As an example, IEs are implemented in the domain of an AS whose number is \textbf{x}. By this same figure, it is observed that the IEs functionally important for the inter-domain operations reside into the upper layers.

It is observed in the implementation model that a classification of relevance is the intensity of aggregation that an IE has in relation to the \textbf{self-*} properties. If an IE, however, has some self-organizing capability, it must participate directly linked to the IE Controller. Even if it participate in the layer of Auxiliary IEs there may be a new IE Controller that logically builds a new layer architecture. 

On the other hand, the representation of the model is logical (abstraction of the physical implementation). Physically, locating an IE in the domain environment is essential. The best alternative is IP addressing, preferably IPv6, for reasons of availability. The IE Controller must maintain a table associating the logic reference with the IP designated by the IE Controller itself, from the premise that an IPv6 block must be available at the beginning of the implementation. However, this is not a fundamental issue, because as will be seen in next section, in the name of security an IP relation as the IE ID will be available in a primitive Domain Name System (DNS), the hosts file allocated internally and with direct link to the IE Controller.


\section{A2RD Use Case}

\begin{figure}[ht]
\centering
\includegraphics[width=.23\textwidth]{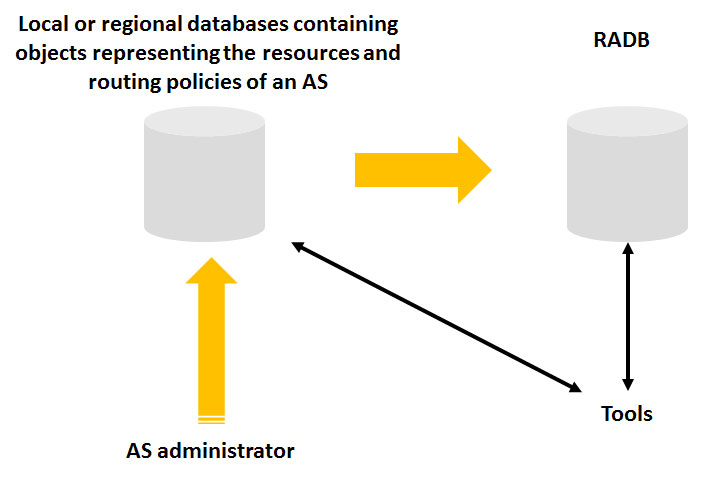}
\includegraphics[width=.23\textwidth]{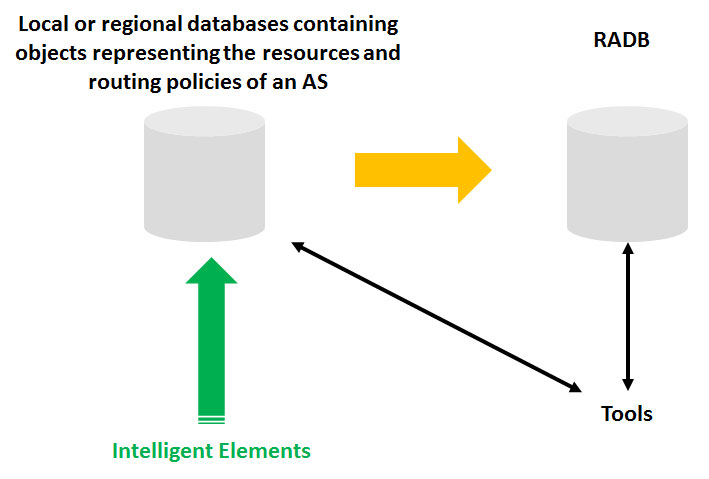}
\caption{IRR model. Objects created by human beings (L) and objects created by IEs (R)}
\label{fig:IRRmodell_EN}
\end{figure}

The A2Rd implementation model has been tested on an Internet Routing Registry\footnote{http://www.irrd.net/} (IRR) server. IRR is an important data base for AS Administrators. But, as the IRR depends on the systematic update of the human being it becomes an unreliable database. By eliminating human intervention and letting the IEs update, the IRR becomes, surprisingly, a reliable database. Figure \ref{fig:IRRmodell_EN} shows that there is no logical difference between human intervention or not.

Numerous IEs have been created, some specialized in the creation of objects, others in the systematic updating of objects and others specialized in capturing information in the various components that makes the AS operational.

\section{Model for knowledge acquisition by IEs}

This experience with de A2RD use case has led to the need to make IEs more autonomous. So, the knowledge required for IEs is available on unstructured bases, in particular, on those produced by the IETF Working Groups (WGs). But they are present in other documents, unofficial and related to equipment and techniques associated with the resources and facilities available in the field of ASes. From this set of available bases, one of the most important is that maintained by the RFC Editor\footnote{\url{https://www.rfc-editor.org/}} represented by more than 8,000 documents called Request for Comments (RFCs).

\begin{figure*}[ht]
\centering
\includegraphics[width=.9\textwidth]{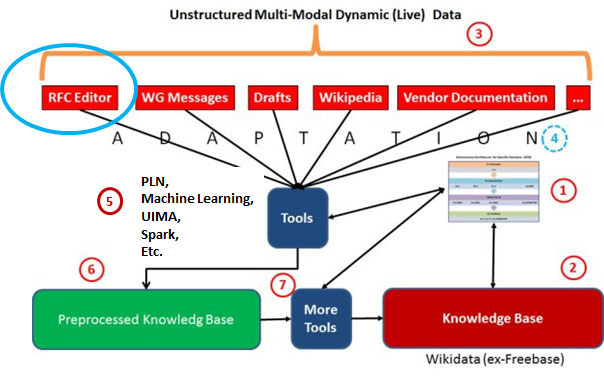}
\caption{Global knowledge capture model for IEs in a restricted domain. Source: \cite{Braga:2015a}}
\label{fig:tese-modeloprincipal_EN}
\end{figure*}

Figure \ref{fig:tese-modeloprincipal_EN} shows the whole process of acquiring knowledge from the unstructured bases (3) through an adaptation in (4) so that they can be submitted to appropriate tools and techniques in (5), as IBM Watson techniques \cite{ferrucci2004building, ferrucci2010building}, whose preliminary result is stored in an intermediate knowledge base (6), which when suffering interference from other more elaborate tools (7) and from the IEs themselves (1), will give rise to an appropriate knowledge base (2).

\section{Research questions involved in the process of knowledge acquisition}

The RFCs repository is constantly updated. Any effective use of this repository implies a continuous and permanent process of manipulation, involving four steps, three of which represent research questions that must be studied.

\begin{enumerate}
\item Capture of the RFC Editor repository in a systematic and continuous way.
\item Lexical refinement of RFCs (\textbf{Research question 1})
\item Semantic distillation and construction of the knowledge base. (\textbf{Research question 2})
\item Knowledge base used by IEs. (\textbf{Research question 3})
\end{enumerate}

\begin{figure}[!htb]
\centering
\includegraphics[width=.48\textwidth]{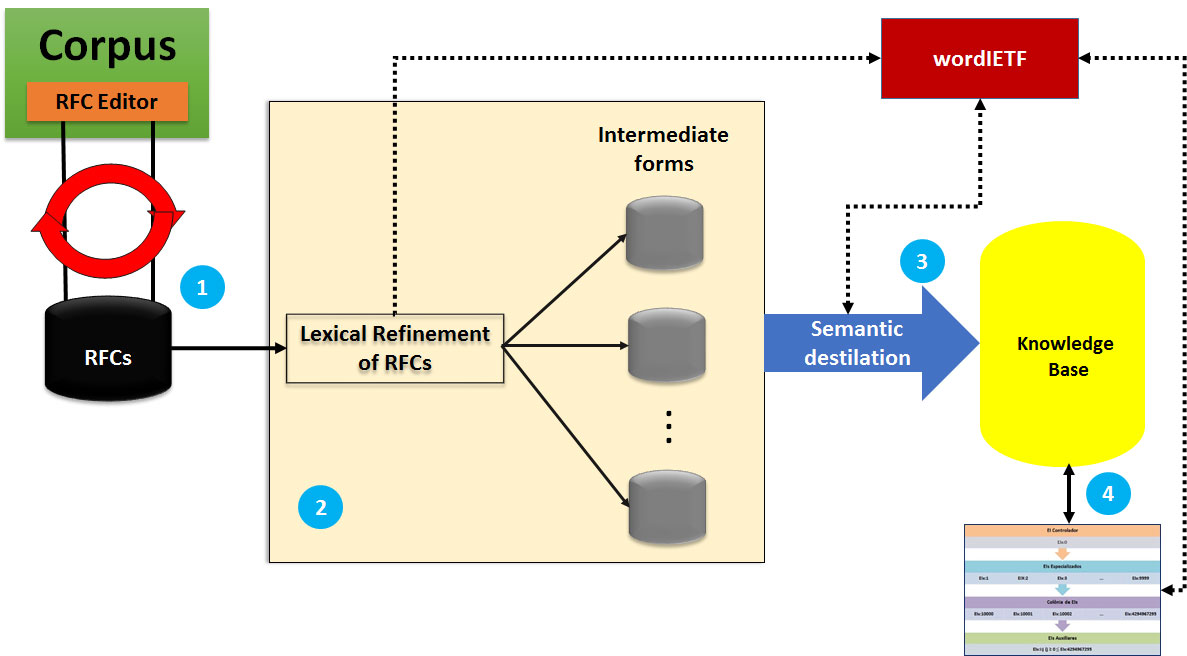}
\caption{Characterization of the problem and its research questions. Source: \cite{Braga:2015a}}
\label{fig:ModeloGlobalQuestoesPesquisa_EN}
\end{figure}

These steps are represented and identified in Figure \ref{fig:ModeloGlobalQuestoesPesquisa_EN}.

\subsection{WordIETF}

In step 2, which identifies research question 1, one has an immediate result. This is a lexical and syntactical  base, called \mbox{\textbf{WordIETF}} with a lot of similarity to WordNet \cite{fellbaum1998wordnet}. \textbf{WordIETF} is built on $\quad$eXtensible Markup Language\footnote{https://www.w3.org/TR/REC-xml/} (XML) and takes full advantage of the building, by the RFC Editor, the  RFC's repository in XML \footnote{https://www.rfc-editor.org/rfc-index.xml}.

The construction of the \textbf{WordIETF} has been tested a series of techniques and recommendations produced by various authors. "Computing is a Natural Science", as said Peter Denning \cite{Denning:2007:CNS:1272516.1272529}.
Semantics is essential for living organisms and defines the relationship between the mind and the world \cite{dodig2008semantics}. 

More recently, authors have admitted several techniques using computational morphology and there is a belief that the contribution of a database such as \textbf{WordIETF} will be fundamental for the evolution of research questions 2 and 3 \cite{dodig2016nature, dodig2011dialogue, dodig2016information}.

\section{Conclusions}

There are many challenges to the progress of the project, among them is the construction of the \textbf{WordIETF} that can meet the demand for construction of ontologies in the area of Internet Infrastructure. The \textbf{WordIETF} development should be cooperative and although there are initiatives of researchers in the area of Internet of Things (IoT) \cite{Hachem:2011}, and by those interested in the Internetware paradigm of the Context Aware Supporting Environment for Internetware\footnote{https://code.google.com/p/casei/} (CASEi), the authors find to have a central coordination, such as the Internet Research Task Force (IRTF), through a multi-stakeholder group.

Without exhausting other proposals it is necessary to develop methodologies for the construction of IEs, in this case, identified as \textbf{Intelligent Objects} (IOs). IO is an intelligent element built with quality, reusable and preserving the knowledge of its life cycle, in the context of A2RD model applications, strengthening the security environment proposed by the DTS model and other related models. Construction of IOs, in an adequate and standardized manner induces interdisciplinary cooperation and rapid development of IEs. The challenge would be to use the methodology INTERA \cite{BragaJ:2015}, adapted to the construction of IOs. INTERA is a successful methodology used in Learning Objects.

\section{Authors' Acknowledgments}

This work was conducted during a scholarship supported by the International Cooperation Program CAPES at the University of Saskatchewan, CA. Financed by CAPES – Brazilian Federal Agency for Support and Evaluation of Graduate Education within the Brazil's Ministry of Education.

%
\bibliographystyle{abbrv}
\bibliography{ms}  

\begin{thebibliography}{10}

\bibitem{Braga:2015a}
J.~Braga.
\newblock {Modelo para Implementação de Elementos Inteligentes em Domínios
  Restritos da Infraestrutura da Internet}.
\newblock Master's thesis, Universidade Presbiteriana Mackenzie, São Paulo,
  SP, 8 2015.

\bibitem{BragaJ:2015}
J.~Braga.
\newblock {\em Objetos de Aprendizagem: Metodologia de Desenvolvimento}.
\newblock Editora da UFABC, São Paulo, 1 edition, 2015.

\bibitem{Braga:2015}
J.~Braga, N.~Omar, and L.~Z. Granville.
\newblock Uma proposta para o uso de elementos inteligentes em domínios
  restritos da infraestrutura da internet.
\newblock In {\em Anais CSBC 2015 - WPIETFIRTF}, Recife, Pernambuco, Brasil,
  jul 2015.

\bibitem{Denning:2007:CNS:1272516.1272529}
P.~J. Denning.
\newblock Computing is a natural science.
\newblock {\em Commun. ACM}, 50(7):13--18, July 2007.

\bibitem{dodig2008semantics}
G.~Dodig-Crnkovic.
\newblock Semantics of information as interactive computation.
\newblock In {\em WSPI}, 2008.

\bibitem{dodig2016information}
G.~Dodig-Crnkovic.
\newblock Information, computation, cognition. agency-based hierarchies of
  levels.
\newblock In {\em Fundamental Issues of Artificial Intelligence}, pages
  139--157. Springer, 2016.

\bibitem{dodig2016nature}
G.~Dodig-Crnkovic.
\newblock Nature as a network of morphological infocomputational processes for
  cognitive agents.
\newblock {\em The European Physical Journal Special Topics}, pages 1--15,
  2016.

\bibitem{dodig2011dialogue}
G.~Dodig-Crnkovic and V.~M{\"u}ller.
\newblock A dialogue concerning two world systems: info-computational vs.
  mechanistic.
\newblock {\em Information and computation}, pages 149--184, 2011.

\bibitem{fellbaum1998wordnet}
C.~Fellbaum.
\newblock {\em WordNet: An Electronic Lexical Database}.
\newblock MIT Press, Cambridge, MA, 1998.

\bibitem{ferrucci2010building}
D.~Ferrucci, E.~Brown, J.~Chu-Carroll, J.~Fan, D.~Gondek, A.~A. Kalyanpur,
  A.~Lally, J.~W. Murdock, E.~Nyberg, J.~Prager, et~al.
\newblock {Building Watson: An overview of the DeepQA project}.
\newblock {\em AI magazine}, 31(3):59--79, 2010.

\bibitem{ferrucci2004building}
D.~Ferrucci and A.~Lally.
\newblock Building an example application with the unstructured information
  management architecture.
\newblock {\em IBM Systems Journal}, 43(3):455--475, 2004.

\bibitem{Hachem:2011}
S.~Hachem, T.~Teixeira, and V.~Issarny.
\newblock Ontologies for the internet of things.
\newblock In {\em Proceedings of the 8th Middleware Doctoral Symposium},
  page~3. ACM, 2011.

\bibitem{Hawkinson:1996}
J.~Hawkinson and T.~Bates.
\newblock {Report on MD5 Performance }.
\newblock Technical report, RFC Editor, March 1996.
\newblock RFC1930. \url{https://tools.ietf.org/rfc/rfc1930.txt}.

\bibitem{Horn:2001}
P.~Horn.
\newblock {Autonomic computing: IBM's Perspective on the State of Information
  Technology}.
\newblock 2001.

\bibitem{Insaurralde:2015}
C.~C. Insaurralde and E.~Vassev.
\newblock Autonomic computing software for autonomous space vehicles.
\newblock In {\em Nature of Computation and Communication}, pages 33--41.
  Springer, 2015.

\bibitem{Mitchell:2013}
J.~Mitchell.
\newblock {Autonomous System (AS) Reservation for Private Use}.
\newblock Technical report, RFC Editor, July 2013.
\newblock RFC6996. \url{https://tools.ietf.org/rfc/rfc6996.txt}.

\bibitem{Movahedi:2012}
Z.~Movahedi, M.~Ayari, R.~Langar, and G.~Pujolle.
\newblock A survey of autonomic network architectures and evaluation criteria.
\newblock {\em Communications Surveys \& Tutorials, IEEE}, 14(2):464--490,
  2012.

\bibitem{Nadeau:2013}
T.~D. Nadeau and K.~Gray.
\newblock {\em SDN: Software Defined Networks}.
\newblock O'Reilly, USA, 1 edition, 2013.

\bibitem{Saltzer:1984}
J.~H. Saltzer, D.~P. Reed, and D.~D. Clark.
\newblock End-to-end arguments in system design.
\newblock {\em ACM Transactions on Computer Systems (TOCS)}, 2(4):277--288,
  1984.

\bibitem{Shukla:2014}
V.~Shukla.
\newblock {\em Introduction to Software Defined Networking}.
\newblock Amazon, USA, 1 edition, 2014.

\bibitem{Sterritt:2005b}
R.~Sterritt and M.~Hinchey.
\newblock Engineering ultimate self-protection in autonomic agents for space
  exploration missions.
\newblock In {\em Engineering of Computer-Based Systems, 2005. ECBS'05. 12th
  IEEE International Conference and Workshops on the}, pages 506--511. IEEE,
  2005.

\bibitem{Wickboldt:2015}
J.~A. Wickboldt, W.~P. De~Jesus, P.~H. Isolani, C.~Bonato~Both, J.~Rochol, and
  L.~Zambenedetti~Granville.
\newblock Software-defined networking: management requirements and challenges.
\newblock {\em Communications Magazine, IEEE}, 53(1):278--285, 2015.

\end{thebibliography}
%
%

\end{document}